\documentclass[letterpaper, 10 pt, conference]{ieeeconf}  

\usepackage{xcolor}
\usepackage{cite} 

\usepackage{booktabs}
\usepackage{multirow}
\usepackage{placeins}
\FloatBarrier

\usepackage{amsmath}
\usepackage{amssymb}

\usepackage{graphicx}
\usepackage{multirow}

\usepackage{array}
\usepackage{caption}
\usepackage{float}
\usepackage{tabularx} 
\usepackage{CJKutf8}
\usepackage{makecell}

\usepackage{colortbl}
\usepackage{booktabs}

\usepackage{url} 

\definecolor{myblue}{RGB}{0,102,204}
\definecolor{darkgreen}{rgb}{0.0, 0.39, 0.0}  
\definecolor{darkred}{rgb}{0.55, 0.0, 0.0}    

\makeatletter
\let\NAT@parse\undefined
\makeatother

\usepackage[colorlinks=true, linkcolor=myblue, citecolor=myblue, urlcolor=myblue]{hyperref}

\IEEEoverridecommandlockouts                              

\title{\LARGE \bf
RoboSpan: Learning Modular Coordination Policies for High-DoF Whole-Body Mobile Manipulation with Teleoperation
}

\author{Keqiang Ren$^{*,1}$, Xuecheng Ye$^{*,1}$,Hanyu Liu$^{*,1}$, Zhihua Wang$^{1}$, Tao Jian$^{1}$, Yunsheng Ma$^{1}$, \\ Jiayi Wen$^{1}$, Xinran Li$^{1}$,Zhigong Song$^{1}$ 
\thanks{$^{*}$These authors contributed equally to this work.}
\thanks{Corresponding author: Zhigong Song ({song\_jnu}@jiangnan.edu.cn).} %
\thanks{
$^{1}$Jiangsu Key Laboratory of Advanced Food Manufacturing Equipment and Technology, School of Mechanical Engineering, Jiangnan University, Wuxi 214122, China. This work was supported by the Wuxi Taihu Lake Talent Plan. This work was also supported by Linkerbot Beijing Technology Co., Ltd. 
}
}

\begin{document}

\maketitle
\thispagestyle{empty}
\pagestyle{empty}


\begin{abstract}
Whole-body mobile manipulation suffers from two fundamental challenges: inherent motion heterogeneity across locomotion, arm, and hand modules, and progressive error accumulation over long-horizon multi-stage tasks. Existing teleoperation platforms also lack integrated dexterous tactile sensing to support coordinated whole-body decision-making. This paper presents RoboSpan, a high-DoF whole-body teleoperation and demonstration platform, and the Hierarchical Modular Framework (HMF) for learning modular coordination policies. HMF functionally decouples the 29-DoF action space into fine hand manipulation, arm pose adjustment, and mobile-base spatial configuration branches, and introduces the Coordinated Modular Embodied Transformer (COMET) with inter-modular cross-attention to enable dynamic information exchange and action coordination. Vision-language model driven hierarchical task decomposition structurally suppresses temporal error propagation, while a full-palm 2D tactile representation encodes distributed contact signals for cross-module tactile empowerment. Experiments on a physical 29-DoF robot demonstrate that the proposed method achieves over 20\% higher task success rate than mainstream baselines across a range of long-horizon whole-body tasks, markedly improving stability and robustness in contact-rich manipulation.

\end{abstract}

\section{INTRODUCTION}






Imitation learning, which acquires manipulation skills from human demonstrations, offers a core route for robots to rapidly learn complex interactive capabilities ~\cite{RN1,RN2}. Driven by the demands of unstructured real-world scenes, robots are increasingly moving beyond fixed workbenches toward whole-body mobile manipulation systems that integrate mobile navigation, bimanual manipulation, and multimodal perception ~\cite{RN3,RN4,RN5}. This evolution also raises two core challenges: motion coordination among heterogeneous body modules and error accumulation in long-horizon complex tasks.
A whole-body mobile manipulation robot typically comprises a mobile base, manipulators, and dexterous hands, which differ markedly in action-space dimensionality, motion scale, control period, and optimization objective ~\cite{RN6,RN7,RN8}: the base handles large-range locomotion and global pose adjustment with a long control period, whereas the arms and hands perform close-range, high-precision fine manipulation ~\cite{RN5,RN6}. Modeling all modules jointly within a single policy sharply expands action dimensionality, increasing learning difficulty while masking the motion characteristics of each module and degrading per-module control precision ~\cite{RN9}. Realizing efficient inter-module information exchange and coordination while preserving each module's specialized control capability is thus a central problem for whole-body embodied intelligence, since forcing heterogeneous modules into one shared output space inevitably trades global mobility for local dexterity and ultimately limits real-world deployment.
Expanding the workspace also complicates task workflows and lengthens execution time. Methods designed for short-horizon, single-scene tasks are prone to cumulative error over multi-stage, long-cycle real-world tasks, causing a steep decline in the middle and late stages ~\cite{RN10}. Vision-language models (VLMs) can decompose high-level natural-language instructions into executable sub-task sequences through semantic understanding ~\cite{RN11,RN12}, offering new possibilities for hierarchical long-horizon execution. However, combining hierarchical task decomposition with low-level body-module coordination into a complete pipeline from semantic instructions to whole-body actions still lacks a systematic solution, which is the gap this work aims to close.
At the manipulation level, existing platforms mostly employ simple grippers or suction cups inadequate for complex dexterous interactions ~\cite{RN13,RN14}. Bionic dexterous hands add DoFs and manipulation flexibility, yet without contact-state feedback they still suffer from unstable grasping and poor precision under vision-limited conditions. Distributed tactile sensing captures contact-force distribution and slip trends that vision cannot, offering an important route to improving fine manipulation ~\cite{RN15,RN16,RN17} Most tactile studies, however, target hand manipulation itself; how tactile information propagates across modules to assist the coordinated decisions of the arms and the mobile base remains largely unexplored.
To address these issues, this paper proposes HMF, a hierarchical modular learning framework for high-DoF whole-body mobile manipulation, together with the companion teleoperation and demonstration platform RoboSpan. The main contributions of this paper are as follows:
\begin{figure*}[htbp]
   \centering
    \includegraphics[width=\textwidth]{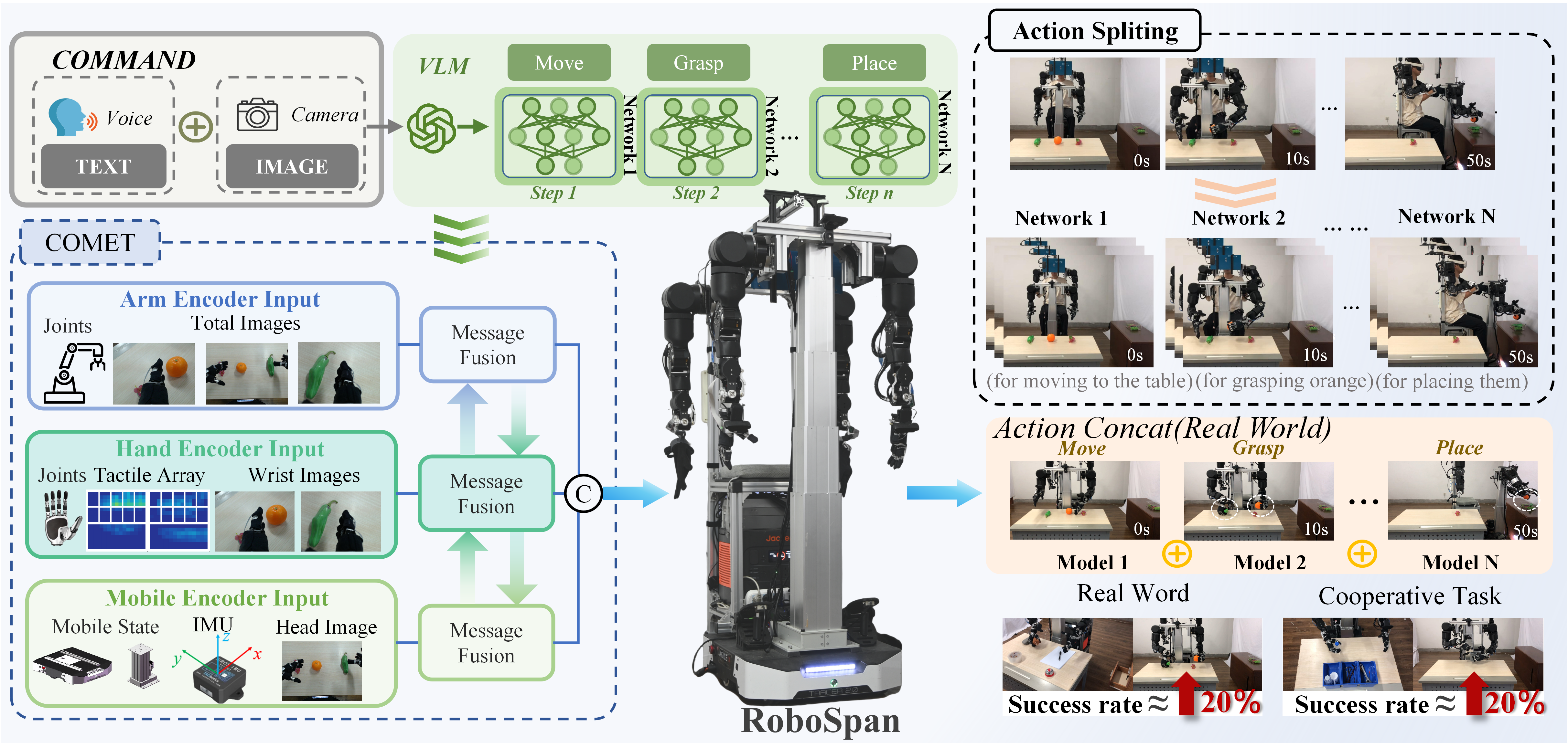} 
    \captionsetup{singlelinecheck=off, justification=raggedright}
    \caption{Overview of the RoboSpan framework for high-DoF whole-body mobile manipulation. RoboSpan integrates whole-body teleoperation, multimodal perception, and modular policy learning, enabling VLM-guided hierarchical task decomposition and coordinated Hand--Arm--Mobile execution for long-horizon manipulation.}
    \label{Overview_RoboSpan}
\end{figure*}

1) \textbf{A high-DoF whole-body teleoperation and demonstration-collection platform:}: RoboSpan integrates mobility, manipulation, and multimodal perception with 29 control DoFs in total, and a foot-IMU-based locomotion/elevation interface enables synchronous multi-module teleoperation and time-aligned data collection.

2) \textbf{Full-palm distributed tactile sensing and 2-D representation:}  a full-palm sensing scheme with a 2-D image representation preserves the  distribution of contact pressure and yields a tactile--visual isomorphic representation, providing a unified feature basis for cross-module tactile empowerment.

3) \textbf{The HMF framework and COMET policy:} HMF suppresses long-sequence error accumulation through VLM-driven stage decomposition, while COMET uses inter-modular cross-attention to improve whole-body coordination and task success while preserving per-module control precision.
\section{RELATED WORK}
\subsection{Teleoperation and Demonstration Collection for Whole-Body Mobile Manipulation}
Teleoperation is a core approach for acquiring high-quality human demonstrations and a key data-collection means for imitation learning ~\cite{RN18,RN19}. As robotic tasks expand from fixed workbenches to large spaces, teleoperation systems have evolved from single-manipulator control toward coordinated control of multiple whole-body modules.
Early work targeted fixed-base upper-limb manipulation, mapping arm and hand motion through motion capture, exoskeletons, or VR interfaces. Open-TeleVision ~\cite{RN18} and OPEN TEACH ~\cite{RN19} built immersive VR-based bimanual systems for upper-limb fine manipulation, while Dexora [23] combined a vision-language model with an exoskeleton interface for more natural dexterous-hand teleoperation. These systems excel at desktop fine manipulation but are constrained by a fixed base and cannot support cross-space mobile manipulation.
To expand the robot workspace, mobile bases were introduced into teleoperation systems. Mobile ALOHA ~\cite{RN20} realized bimanual mobile-manipulation control with leader--follower arms and a movable base; BEHAVIOR Robot Suite ~\cite{RN21} combined dual manipulators with a mobile base for everyday household tasks; and RoboMatch ~\cite{RN12} proposed a cockpit-style architecture supporting continuous long-horizon demonstration collection. As Table \ref{tab:teleoperation_comparison} summarizes, however, these platforms mostly use simple grippers or suction cups and lack high-DoF dexterous hands and distributed tactile sensing; their locomotion and upper-limb channels are also mutually independent, making it hard to adjust the base while sustaining continuous bimanual fine manipulation, which compromises multi-module demonstration synchronization.
\begin{table*}[htbp]
\centering
\caption{COMPARISON OF EXISTING ROBOT TELEOPERATION SYSTEMS}
\label{tab:teleoperation_comparison}
\resizebox{\textwidth}{!}{
\begin{tabular}{l c c c c c c}
\toprule
\makecell{\textbf{Teleop System}} & 
\makecell{\textbf{Interaction Interface}} & 
\makecell{\textbf{Bimanual}} & 
\makecell{\textbf{D-Handl}} & 
\makecell{\textbf{Elevation}} & 
\makecell{\textbf{Whole-Body} \\ \textbf{Teleop}} & 
\makecell{\textbf{Wild}} \\
\midrule
AnyTeleop~\cite{RN22} & Vision-based teleoperation &
\textcolor{darkred}{$\times$} & \textcolor{darkgreen}{$\checkmark$} & \textcolor{darkred}{$\times$} & \textcolor{darkred}{$\times$} & \textcolor{darkred}{$\times$} \\
Open-TeleVision~\cite{RN18} & VR Arm-Hand Teleop &
\textcolor{darkgreen}{$\checkmark$} & 
\textcolor{darkgreen}{$\checkmark$} & \textcolor{darkred}{$\times$} & \textcolor{darkred}{$\times$} & \textcolor{darkred}{$\times$} \\
OPEN TEACH~\cite{RN19} & VR Arm-Hand Teleop &
\textcolor{darkgreen}{$\checkmark$} & \textcolor{darkgreen}{$\checkmark$} &
\textcolor{darkred}{$\times$} & \textcolor{darkred}{$\times$} &
\textcolor{darkred}{$\times$} \\
Mobile ALOHA~\cite{RN20} & Leader arms + backdrivable base &
\textcolor{darkgreen}{$\checkmark$} & \textcolor{darkred}{$\times$} & \textcolor{darkgreen}{$\checkmark$} &
\textcolor{darkgreen}{$\checkmark$} & \textcolor{darkred}{$\times$} \\
BEHAVIOR Robot Suite~\cite{RN21} & Twin arms + controllers &
\textcolor{darkgreen}{$\checkmark$} & \textcolor{darkred}{$\times$} & \textcolor{darkred}{$\times$} &
\textcolor{darkgreen}{$\checkmark$} & \textcolor{darkred}{$\times$} \\
Dexora~\cite{RN23} & VR Hand Teleop + Exo-Teleop &
\textcolor{darkgreen}{$\checkmark$} & \textcolor{darkgreen}{$\checkmark$} & \textcolor{darkred}{$\times$} &
\textcolor{darkred}{$\times$} & \textcolor{darkred}{$\times$} \\
RoboMatch~\cite{RN12} & Cockpit-style teleoperation &
\textcolor{darkgreen}{$\checkmark$} & \textcolor{darkred}{$\times$} & \textcolor{darkgreen}{$\checkmark$} &
\textcolor{darkgreen}{$\checkmark$} & \textcolor{darkgreen}{$\checkmark$} \\
\textbf{RoboSpan (ours)} & Leader arms + gloves + pedals &
\textcolor{darkgreen}{$\checkmark$} & \textcolor{darkgreen}{$\checkmark$} & \textcolor{darkgreen}{$\checkmark$} &
\textcolor{darkgreen}{$\checkmark$} & \textcolor{darkgreen}{$\checkmark$} \\
\bottomrule
\end{tabular}
}
\end{table*}
\subsection{Distributed Tactile Sensing for Dexterous Manipulation}
Tactile sensing is a core technology for obtaining contact-state information and improving fine-manipulation robustness ~\cite{RN24,RN25}. In complex interactions, vision alone cannot capture contact-force distribution, object slippage, or local deformation, and fusing vision with touch can effectively improve grasping stability and manipulation precision ~\cite{RN26,RN27}.
Tactile hardware and algorithms have both advanced: flexible tactile arrays and electronic skins enable multi-region contact-force detection on dexterous-hand surfaces ~\cite{RN14}, while ViTacFormer ~\cite{RN27} and 3D-ViTac ~\cite{RN28} learn visuo-tactile feature fusion that improves contact-rich performance and TACT ~\cite{RN16} introduced the tactile modality into whole-body humanoid control.
Most existing work, however, treats touch as a local signal for hand-centric tasks such as grasping and in-hand manipulation ~\cite{RN29,RN30}. In whole-body mobile systems its value remains underexploited: contact states can also constrain arm-pose adjustment and base-position calibration, and how tactile information propagates across modules to drive coordinated whole-body decisions still lacks systematic study.
\subsection{Modular Imitation Learning and Long-Horizon Task Policies}
Imitation learning is the dominant paradigm for robots to acquire complex manipulation skills from human demonstrations ~\cite{RN31,RN32}. Transformer- and diffusion-based policy-learning methods have markedly improved performance and generalization on complex manipulation tasks ~\cite{RN33,RN34}, and follow-up work has further extended policy learning through latent action representation, vision-language perception, and consistency distillation to enhance deployment efficiency and fine-manipulation capability across scenarios ~\cite{RN35,RN36,RN37,RN38}.
\begin{figure*}[htbp]
    \centering
    \includegraphics[width=\textwidth]{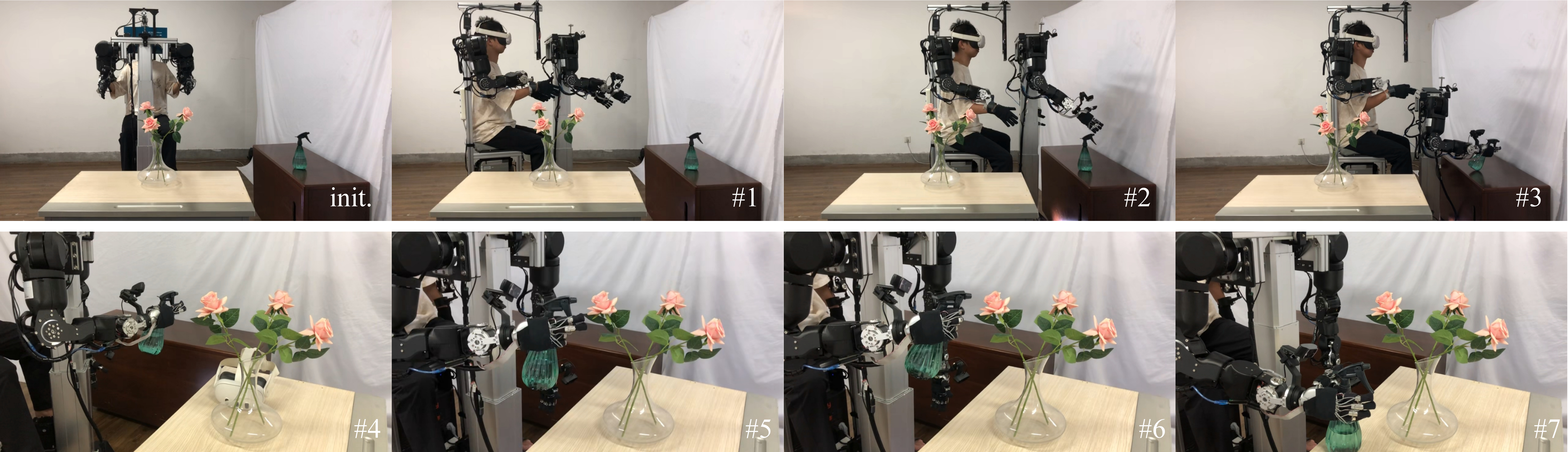} 
    \captionsetup{singlelinecheck=off, justification=raggedright}
    \caption{RoboSpan teleoperation demonstration.}
    \label{TeleOperation_real}
\end{figure*}
\begin{figure}[htbp]
    \centering
    \includegraphics[width=\columnwidth]{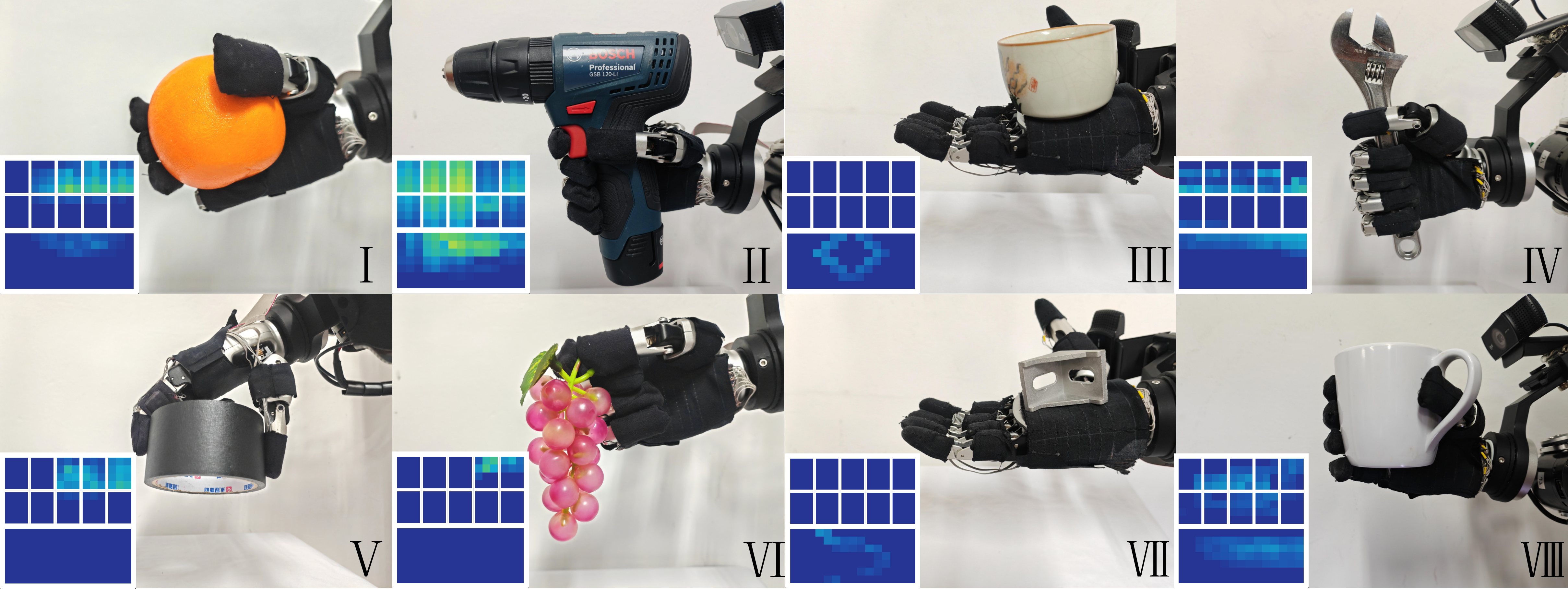} 
    \captionsetup{singlelinecheck=off, justification=raggedright}
    \caption{The dexterous hand and full-palm distributed tactile sensing across diverse objects (I--VIII), together with the corresponding  tactile maps.}
    \label{Tactical} 
\end{figure}
For whole-body mobile manipulation, existing studies follow two main routes. Unified policy modeling integrates the action spaces of all body modules into a single output ~\cite{RN6,RN9}, but sharply expands action dimensionality and makes it hard to balance the motion characteristics and control precision of different modules. Modular decomposition instead designs independent controllers per body module ~\cite{RN7,RN8}, reducing learning difficulty but lacking dynamic inter-module information exchange and thus whole-body coordination in complex tasks. Most modular studies target single-module manipulation in a fixed workspace, such as standalone manipulators and end-effectors ~\cite{RN39,RN40,RN41}; for example, InterACT ~\cite{RN41} introduces hierarchical attention for action chunking in bimanual manipulation, yet remains limited to fixed-base settings without whole-body mobility.
Long-horizon multi-stage tasks pose another major challenge for imitation learning, as conventional single-step policies accumulate error over long cycles and degrade substantially toward the end of a task ~\cite{RN10}. Real-world mobile manipulation combines spatial locomotion, multi-stage manipulation, and long-horizon decision-making, which conventional desktop policies cannot directly handle ~\cite{RN42}. Relay Policy Learning ~\cite{RN10} mitigates this through policy chaining but relies on manual stage partitioning, while VLMs enable automatic semantic decomposition into sub-task sequences ~\cite{RN11,RN12}, offering a new route for long-horizon execution. Existing studies, however, largely separate task decomposition from low-level control policies and fail to integrate hierarchical task structure with modular body coordination, leaving no complete pipeline from high-level semantic instructions to low-level whole-body actions.

\section{METHOD}

\subsection{RoboSpan: A Whole-Body Teleoperation and Demonstration-Collection Platform} 





Existing mobile-manipulation teleoperation systems generally lack sufficient end-effector DoFs, exhibit weak coupling between locomotion and upper-limb control, and suffer from insufficient multimodal synchronization, limiting the collection of high-DoF whole-body coordinated demonstrations. To address these limitations, we build RoboSpan, an integrated platform consisting of a robot body and a teleoperation cockpit. It synchronously controls a mobile base (Tracer 2.0), dual 7-DoF OpenArmX manipulators, dual 6-DoF Linker Hand O6 dexterous hands, and a lifting mechanism, while collecting time-aligned multimodal demonstrations covering vision, touch, and proprioception.

The robot body integrates four functionally heterogeneous actuation modules—dual manipulators, dual dexterous hands, a mobile base, and an active lifting mechanism—with 29 control DoFs in total, enabling multi-scale operations ranging from floor-level grasping and desktop manipulation to high-place object placement. The manipulators perform target localization and end-effector pose adjustment within the workspace, the dexterous hands execute fine grasping and contact-rich interaction, the mobile base provides large-range locomotion and global pose adjustment, and the lifting mechanism extends the reachable workspace across different operating heights.

As shown in Fig.~\ref{TeleOperation_real}, the proposed teleoperation cockpit provides an intuitive interface for whole-body robot control and multimodal demonstration collection. It adopts a master–slave mapping architecture consisting of dual leader arms, hand-control interfaces, and a head-mounted visual-feedback device. To resolve the conflict between locomotion control and upper-limb manipulation over limited input channels, a foot-IMU-based control scheme is introduced to derive base-motion and lifting commands from foot-posture variations, allowing the operator’s hands to remain dedicated to dexterous manipulation while maintaining continuous whole-body coordination. All sensing modalities, including vision, tactile feedback, and proprioception, are synchronized under a unified timestamp, providing temporally aligned demonstrations for subsequent heterogeneous-module coordination learning.

\begin{figure*}[htbp]
    \centering
    \includegraphics[width=\textwidth]{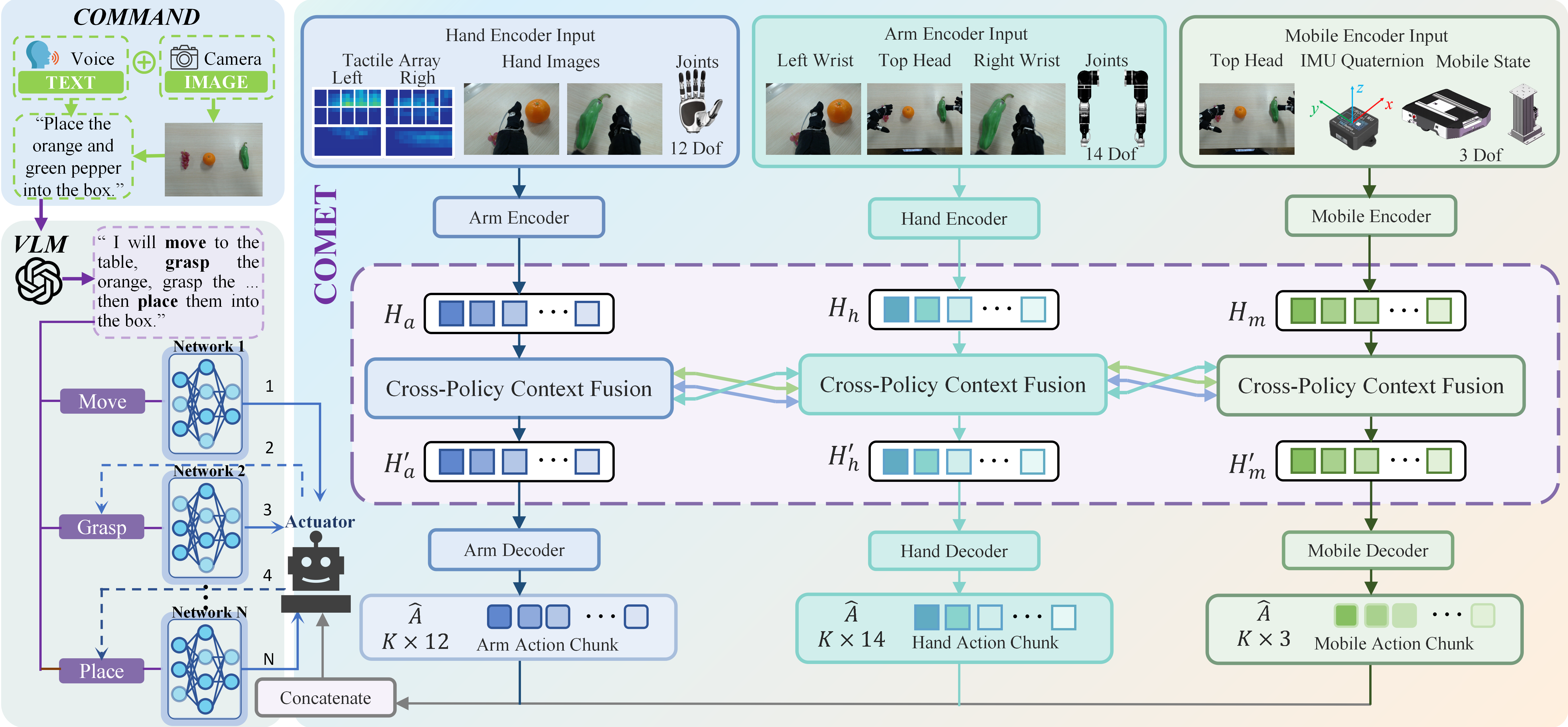} 
    \captionsetup{singlelinecheck=off, justification=raggedright}
    \caption{Overall network architecture of HMF, which consists of three core modules—Hand, Arm, and Mobile—each with a dedicated encoder/decoder and coupled through cross-policy context fusion (IMCA) under VLM-driven multi-stage decomposition. }
    \label{HMF}
\end{figure*}

\subsection{Full-Palm Distributed Tactile Sensing and 2-D Representation}






To overcome the limited coverage of existing tactile schemes and the difficulty
of fusing discrete signals across modules, we build a full-palm distributed
tactile sensing system on the dexterous-hand surface and propose a 2-D
tactile-image representation. On the key contact regions of the Linker Hand O6,
including fingertips, finger pads, and the palm, we deploy 11 modular flexible
tactile units per hand, covering the major interaction regions of the hand. 

As shown in Fig.~\ref{Tactical}, the dexterous hand equipped with the
full-palm distributed tactile sensing system enables rich contact perception
across diverse objects. To make tactile information transferable across modules
in a representation format compatible with visual features, we map discrete
tactile responses into 2-D tactile images. Following
the physical arrangement of sensing units, raw tactile signals are interpolated
into a $15\times15$ two-dimensional tactile matrix, where each coordinate
corresponds to an actual hand-contact region and each pixel value represents
the local tactile response intensity. This representation preserves the 
distribution of contact signals for learning contact locations, pressure
patterns, and dynamic interaction states, while making tactile and visual
information isomorphic in data format, establishing a unified basis for
cross-modal fusion and cross-module interaction.

RoboSpan then performs unified synchronous multimodal acquisition. Besides
tactile images, it records multi-view images, dual-arm and dual-hand joint
states, base and lifting-mechanism states, and teleoperation commands. All
modalities are aligned under a unified timestamp at a sampling rate of 50 Hz.
This produces a complete demonstration representation consisting of visual
observations, tactile images, proprioceptive states, and action commands, while
preserving the temporal correspondence between robot trajectories and
multimodal perception, providing unified data support for the subsequent COMET
network.

\subsection{HMF: A Hierarchical Modular Coordination Learning Framework}

To address heterogeneous-module mismatch and long-horizon error accumulation,
we propose the \textbf{Hierarchical Modular Framework (HMF)}, a two-level
progressive architecture. As shown in Fig.~\ref{HMF}, HMF consists of a
high-level task decomposition module and a low-level coordinated modular policy
module. Its upper level employs a vision-language model (VLM) for semantic
parsing and hierarchical task decomposition, splitting a long-horizon task into
multiple semantically clear sub-task stages to structurally suppress temporal
error accumulation. Its lower level instantiates, for each sub-task stage, the
proposed \textbf{Coordinated Modular Embodied Transformer (COMET)} for dynamic
coordinated control of heterogeneous body modules.

Given a natural-language task instruction, HMF uses the VLM together with
current visual observations to parse the task semantics and execution logic,
partitioning the complex long-horizon goal into $N$ consecutive sub-tasks with
clear operational purposes:

\begin{equation}
\mathcal{T}=\{T_1,T_2,\cdots,T_N\}.
\end{equation}

Each sub-task represents an independent operation stage, such as approaching,
grasping, or placing, with distinct operational goals and degrees of module
involvement. We learn independent policy parameters $\theta_k$ for each stage
to avoid interference caused by cross-stage action-distribution differences.
During execution, the system invokes the corresponding COMET policy according
to the current stage:

\begin{equation}
A_t=\pi_{\theta_k}(O_t),\quad t\in T_k .
\end{equation}
where $O_t$ denotes the robot observation at time $t$, $A_t$ denotes the
predicted action, and $\pi_{\theta_k}$ represents the COMET policy of the
$k$-th task stage. Staged decomposition confines the errors of long sequences
within individual sub-tasks, preventing continuous error accumulation.

For heterogeneous-module coordination within each sub-task, COMET
functionally decouples the unified 29-DoF action space into three independent
modules: \textit{Hand}, \textit{Arm}, and \textit{Mobile}. Each module learns a
specialized policy in its own feature space while achieving whole-body
coordination through dynamic inter-module interaction.

Each module constructs a dedicated observation space according to its
functional role and control objective. The Hand module receives dual-hand joint
states, tactile images, and hand-camera observations for contact interaction and
fine manipulation. The Arm module receives dual-arm joint states, hand-camera
and top-camera observations for end-effector adjustment and workspace
coordination. The Mobile module receives operator motion intent sensed by foot
IMUs, the base/lifting mechanism state, and top-camera observations for
whole-body spatial configuration.

At time $t$, the observation spaces of the three modules are formulated as:

\begin{equation}
O_t^{H}
=
\{q_t^{H},\tau_t^{H},I_t^{hand}\},
\end{equation}

\begin{equation}
O_t^{A}
=
\{q_t^{A},I_t^{hand},I_t^{top}\},
\end{equation}

\begin{equation}
O_t^{M}
=
\{\phi_t^{ped},s_t^{M},I_t^{top}\}.
\end{equation}
where $O_t^{H}$, $O_t^{A}$, and $O_t^{M}$ denote the observations of the Hand,
Arm, and Mobile modules, respectively. $q_t^{H}$ and $q_t^{A}$ represent the
dual-hand and dual-arm joint states, $\tau_t^{H}$ denotes tactile images,
$I_t^{hand}$ and $I_t^{top}$ represent hand-camera and top-camera observations,
and $\phi_t^{ped}$ denotes operator motion intent obtained from IMU pedals.
The state of the mobile base and lifting mechanism is defined as:

\begin{equation}
s_t^{M}=\{v_t,\omega_t,h_t\},
\end{equation}
where $v_t$, $\omega_t$, and $h_t$ represent the base linear velocity, angular
velocity, and lifting height, respectively.

Each body module extracts specialized representations through an independent
Transformer encoder, mapping heterogeneous modules into separate feature
spaces. This design avoids learning heterogeneous motion behaviors in a unified
high-dimensional space and enables each module to learn task-specific
representations:

\begin{equation}
h_t^{i}=E_i(O_t^{i}),
\quad i\in\{H,A,M\}.
\end{equation}
where $E_i$ denotes the encoder of module $i$.

However, fully independent encoding separates the state correlations among
modules and may result in insufficient coordination. Therefore, COMET
introduces an \textbf{Inter-Modular Cross Attention (IMCA)} mechanism, where
each module feature serves as the query and the features from other modules
serve as keys and values for dynamic information selection and fusion:

\begin{equation}
\tilde{h}_t^{i}
=
h_t^{i}
+
\sum_{j\neq i}
\mathrm{CrossAttention}
(h_t^{i},h_t^{j},h_t^{j}),
\quad i,j\in\{H,A,M\}.
\end{equation}

IMCA enables local information propagation among heterogeneous modules. For
example, when the Hand module detects object slippage, tactile information is
transmitted to the Arm module for real-time pose adjustment and further to the
Mobile module for fine base-position correction, achieving contact-driven
whole-body coordination.

After cross-module fusion, each module predicts its future action chunk through
an independent Transformer decoder:

\begin{equation}
A_t^{i}=D_i(\tilde{h}_t^{i}),
\quad i\in\{H,A,M\}.
\end{equation}

The three module outputs are then concatenated according to the robot control
interface:

\begin{equation}
A_t=[A_t^{H},A_t^{A},A_t^{M}].
\end{equation}

For the $k$-th task stage, this produces the action-chunk sequence
$\hat{A}_t^{(k)}$. Once the current sub-task is completed, the system switches
to the next-stage COMET policy until the entire task is completed:

\begin{equation}
\pi_{\theta_1}
\rightarrow
\pi_{\theta_2}
\rightarrow
\cdots
\rightarrow
\pi_{\theta_N}.
\end{equation}

Each stage-specific COMET policy is optimized using a standard
behavior-cloning objective. The overall loss is formulated as a weighted sum
of the three module action prediction errors:

\begin{equation}
\mathcal{L}
=
\sum_{i=1}^{3}
\lambda_i
\|A_i-\hat{A}_i\|_2^2 ,
\end{equation}
where $A_i$ and $\hat{A}_i$ denote the demonstrated and predicted action
chunks of module $i$, respectively.

Through the three-level structure of task-stage decomposition,
module-specific representation, and cross-module dynamic interaction, HMF
jointly addresses long-horizon error accumulation and heterogeneous-module
coordination, enabling efficient policy learning for high-DoF whole-body mobile
manipulation.

\begin{table}[t]
\centering
    \caption{\small Ablation study of task decomposition and modular strategy in the HMF framework (\%)}
\label{two}
\renewcommand{\arraystretch}{1.3}

\begin{tabular*}{\columnwidth}{@{\extracolsep{\fill}}cccccc}
\toprule
Method  & Task 1 & Task 2 & Task 3 & Average \\
\midrule

ACT
 & 48 & 40 & 36 & 41.3 \\

DP U-Net
 & 40 & 28 & 26& 31.3 \\

Modular ACT
 & 42 & 33 & 30& 34.1 \\

COMET
 & 52 & 46 & 42& 46.7 \\

\textbf{HMF (Ours)}
 & \textbf{60} & \textbf{54} & \textbf{50} & \textbf{54.7} \\

\bottomrule
\end{tabular*}

\end{table}

\begin{table*}[t]
\centering
\caption{Stage-wise success rate on whole-body mobile manipulation tasks}
\label{three}
\begin{tabular}{cccccccccccc}
\toprule
\multirow{3}{*}{Method} 
& \multicolumn{6}{c}{\textbf{Desktop Organization (3000 time steps)}} 
& \multicolumn{5}{c}{\textbf{Trash Cleaning (3000 time steps)}} \\
\cmidrule(lr){2-7} \cmidrule(lr){8-12}

& Forward 
& \begin{tabular}{@{}c@{}}Grasp\\Tape\end{tabular}
& \begin{tabular}{@{}c@{}}Grasp\\Bowl\end{tabular}
& Put
& Close
& Back

& \begin{tabular}{@{}c@{}}Grasp\\Cup\end{tabular}
& \begin{tabular}{@{}c@{}}Grasp\\Paper\end{tabular}
& Move
& Lift
& Put \\

\midrule

ACT~\cite{RN33} 
& 100\% & 54\% & 36\% & 18\% & 6\% & 4\%
& 56\% & 44\% & 26\% & 24\% & 10\% \\

DP U-Net~\cite{RN34}
& 100\% & 32\% & 18\% & 6\% & 0\% & 0\%
& 30\% & 14\% & 0\% & 0\% & 0\% \\

\textbf{HMF(Ours)}
& \textbf{100\%} & \textbf{76\%} & \textbf{68\%}
& \textbf{60\%} & \textbf{48\%} & \textbf{46\%}
& \textbf{72\%} & \textbf{68\%} & \textbf{54\%}
& \textbf{52\%} & \textbf{50\%} \\

\bottomrule
\end{tabular}
\end{table*}

\begin{figure*}[htbp]
    \centering
    \includegraphics[width=\textwidth]{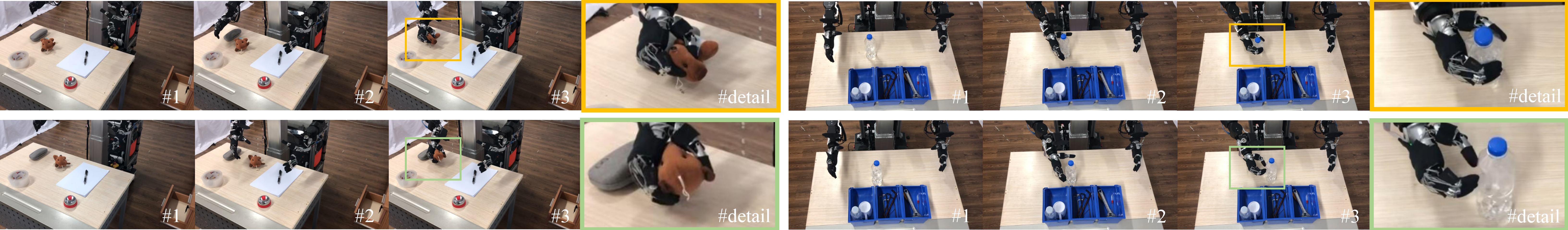} 
    \captionsetup{singlelinecheck=off, justification=raggedright}
    \caption{Qualitative comparison between ACT (top row) and the proposed HMF (bottom row) on real whole-body mobile manipulation tasks; HMF improves performance by referencing the tactile module for perceptual coordination.}
    \label{Experient_Tactical}
\end{figure*}



\begin{table}[t]
\centering
\caption{\small Ablation of distributed tactile sensing and modular coordination in contact-rich manipulation tasks(\%)}
\label{four}
\renewcommand{\arraystretch}{1.3}

\begin{tabular*}{\columnwidth}{@{\extracolsep{\fill}}ccccc}
\toprule
Method & Task 1 & Task 2 & Task 3 & Average \\
\midrule
ACT w/o t
& 40 & 66 & 48 & 51.3 \\
ACT + t
& 44 & 70 & 50 & 54.7 \\
Modular ACT + t
& 36 & 60 & 44 & 46.7 \\
\textbf{COMET (Ours)}
& \textbf{48} & \textbf{76} & \textbf{60} & \textbf{61.3} \\
\bottomrule
\end{tabular*}
\end{table}


\section{EXPERIMENT AND ANALYSIS}


To validate the proposed platform, sensing method, and learning framework, we conduct three groups of real-world experiments on the physical RoboSpan platform, corresponding to three objectives: first, verifying HMF's advantages in heterogeneous-module coordination and long-horizon execution; second, ablating full-palm tactile sensing and the cross-module coordination mechanism on contact-rich tasks; and third, evaluating the platform's demonstration-collection stability and data quality. All policies are trained on human teleoperation demonstrations.

\subsection{Validation of the HMF Framework}


To verify the performance of HMF in heterogeneous-module coordination and long-horizon execution, we select three representative whole-body robot tasks — trash cleaning, fruit transfer, and desktop organization. Each task involves multi-stage continuous operations and cross-module coordination. We adopt the behavior-cloning policy ACT and the diffusion policy DP U-Net as comparison baselines. 
Table \ref{two} presents the average success rates across the three tasks. The unified ACT and DP U-Net achieve poor overall performance. COMET yields moderate improvements via module decomposition, yet the absence of inter-module information exchange limits coordination and performance gains. HMF achieves the highest success rate on all three tasks. With the same module partition as COMET, introducing cross-attention and staged task decomposition improves the average success rate, verifying the effectiveness of the COMET coordination mechanism and hierarchical task decomposition.

To further verify the model’s ability to suppress long-horizon error accumulation, Table \ref{three} reports the per-stage success rates for desktop organization and trash cleaning. The success rates of ACT and DP U-Net drop sharply as the task proceeds. HMF confines errors within individual subtasks, with a gentle decline in success rate and maintains a relatively high success rate at the final stage. These results demonstrate that hierarchical task decomposition can effectively suppress long-sequence error accumulation and markedly improve the stability of whole-body robot long-horizon execution.


\subsection{Ablation of Tactile Sensing and Cross-Module Coordination}

To analyze the roles of full-palm tactile sensing and the cross-module coordination mechanism, we conduct ablation experiments on three contact-sensitive tasks (toy transfer, bell pressing, and transparent-bottle transfer) with four settings: ACT without touch, unified ACT with touch, modular ACT with touch, and the full HMF.

As Table \ref{four} shows, adding tactile information to the unified policy raises the average success rate from 51.3\% to 54.7\%. Distributed touch provides contact-state constraints to improve grasping stability and manipulation precision. Notably, the fully decoupled Modular ACT + t even underperforms ACT + t: module decomposition severs the transmission of tactile information across modules, so hand tactile signals cannot be passed to the arm and mobile modules, weakening coordination and offsetting the performance gains brought by modularization.

In contrast, HMF enables tactile information to propagate dynamically among the three modules via cross-attention. It optimizes hand manipulation while assisting arm-pose adjustment and base-position calibration, achieving an average success rate of 61.3\%. As shown in Fig.~\ref{Experient_Tactical}, without tactile input, the robot frequently fails in contact-sensitive manipulation because it cannot adequately perceive object contact and interaction states; with tactile sensing, the robot achieves more stable grasping and more reliable task execution. This demonstrates that the value of tactile information extends beyond hand manipulation itself. Through cross-module coordination, tactile perception enhances whole-body operation as a whole.



\subsection{Evaluation of the RoboSpan Teleoperation Platform}



To evaluate RoboSpan's demonstration-collection capability and data quality, we use VR teleoperation and exoskeleton teleoperation as baselines and compare them along two dimensions: demonstration-collection stability and policy-training effectiveness.

\begin{figure}[htbp]
    \centering
    \includegraphics[width=\columnwidth]{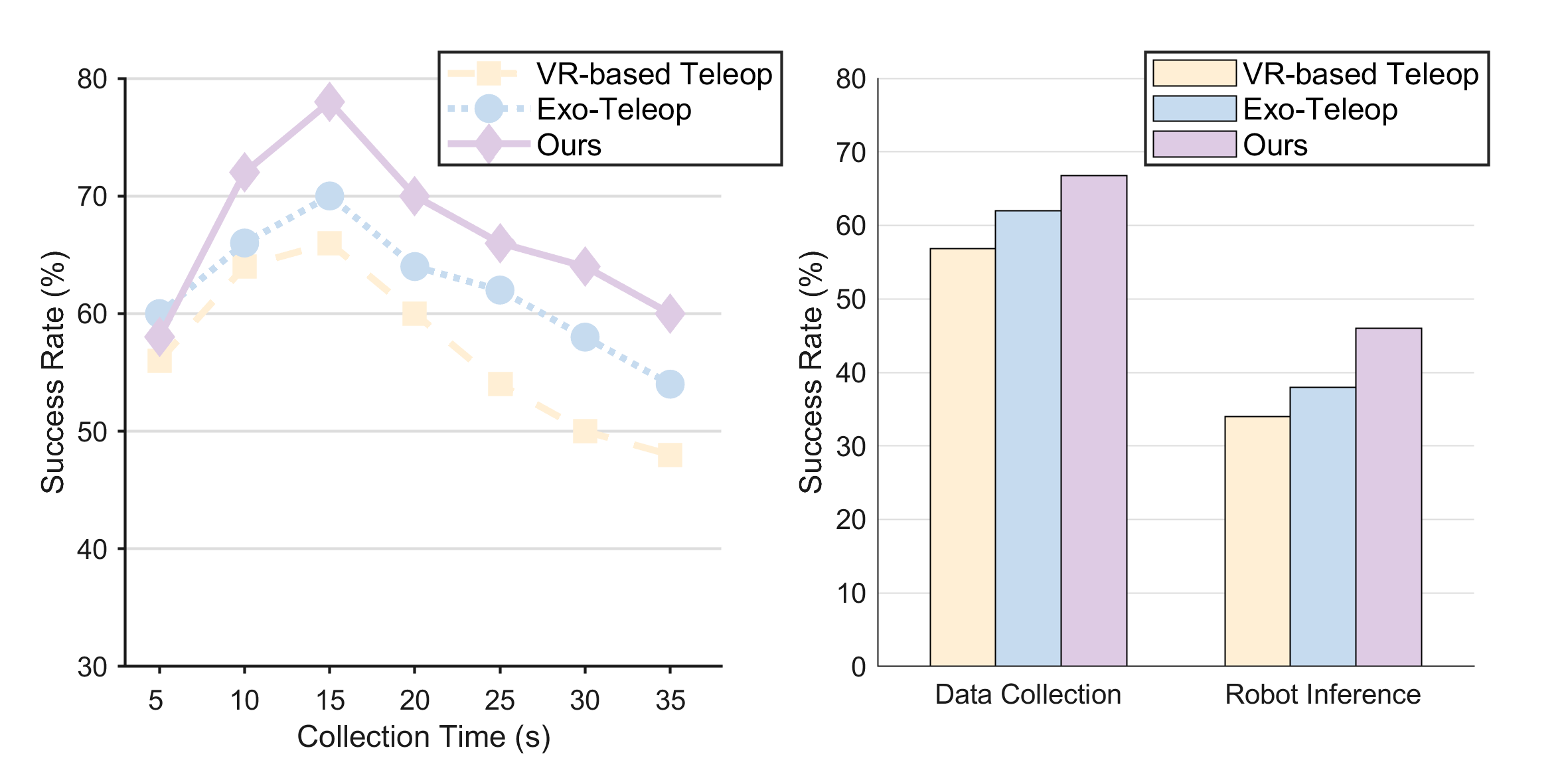} 
    \captionsetup{singlelinecheck=off, justification=raggedright}
    \caption{ Impact of RoboSpan on task performance. (a) Success rate vs. task duration during data collection. (b) Average success rate at the data-collection and robot-inference stages.}
    \label{Experiment_RoboSpan} 
\end{figure}

\hyperref{Fig.~\ref{Experiment_RoboSpan}}(a) shows the one-shot collection success rate under different task durations. As duration increases, all three methods decline, but RoboSpan remains the highest across every range. For long-horizon tasks longer than 20 s, the VR and exoskeleton methods decline markedly, whereas RoboSpan, benefiting from its foot-IMU locomotion scheme, lets the operator adjust the base without interrupting hand operations, preserving long-horizon continuity and synchronization for a more pronounced stability advantage.

\hyperref{Fig.~\ref{Experiment_RoboSpan}}(b) compares the average success rates at the data-collection and policy-inference stages. Policies trained on RoboSpan-collected data achieve the highest real-task success, improving by about 15\% and 10\% over the VR and exoskeleton schemes, respectively. This indicates that the integrated whole-body platform provides more synchronized and complete demonstrations, effectively improving the final performance of imitation-learning policies.

\section{CONCLUSION}


This paper presents RoboSpan, an integrated teleoperation and demonstration-collection platform for high-DoF whole-body mobile manipulation that enables multimodal whole-body demonstration data capture and precise real-world manipulation execution. We further propose HMF, a hierarchical modular coordinated imitation learning framework for long-horizon whole-body mobile manipulation. This VLM-driven framework decomposes task stages to structurally reduce error accumulation over long execution horizons, and leverages COMET with IMCA to achieve dynamic coordination across heterogeneous body modules, with full-palm tactile representations empowering contact-aware whole-body decision-making. Real-robot experiments verify over 20\% average success improvement over mainstream baselines and notable robustness gains in contact-rich tasks. Future work will explore smoother subtask transitions within HMF to enable more cohesive inference, more efficient inter-module information exchange, and policy generalization in open scenes.

\bibliographystyle{IEEEtran}

\begin{thebibliography}{10}
\providecommand{\url}[1]{#1}
\csname url@samestyle\endcsname
\providecommand{\newblock}{\relax}
\providecommand{\bibinfo}[2]{#2}
\providecommand{\BIBentrySTDinterwordspacing}{\spaceskip=0pt\relax}
\providecommand{\BIBentryALTinterwordstretchfactor}{4}
\providecommand{\BIBentryALTinterwordspacing}{\spaceskip=\fontdimen2\font plus
\BIBentryALTinterwordstretchfactor\fontdimen3\font minus \fontdimen4\font\relax}
\providecommand{\BIBforeignlanguage}[2]{{%
\expandafter\ifx\csname l@#1\endcsname\relax
\typeout{** WARNING: IEEEtran.bst: No hyphenation pattern has been}%
\typeout{** loaded for the language `#1'. Using the pattern for}%
\typeout{** the default language instead.}%
\else
\language=\csname l@#1\endcsname
\fi
#2}}
\providecommand{\BIBdecl}{\relax}
\BIBdecl

\bibitem{RN1}
B.~D. Argall, S.~Chernova, M.~Veloso, and B.~Browning, ``{A Survey of Robot Learning from Demonstration},'' \emph{Robot. Auton. Syst.}, vol.~57, no.~5, pp. 469--483, 2009.

\bibitem{RN2}
T.~Osa, J.~Pajarinen, G.~Neumann, J.~A. Bagnell, P.~Abbeel, and J.~Peters, ``{An Algorithmic Perspective on Imitation Learning},'' \emph{Found. Trends Robot.}, vol.~7, no. 1--2, pp. 1--179, 2018.

\bibitem{RN3}
M.~Mittal, D.~Hoeller, F.~Farshidian, M.~Hutter, and A.~Garg, ``{Articulated Object Interaction in Unknown Scenes with Whole-Body Mobile Manipulation},'' \emph{Proc. IEEE/RSJ Int. Conf. Intelligent Robots and Systems (IROS)}, pp. 1647--1654, 2022.

\bibitem{RN4}
Y.~Jiang, R.~Zhang, J.~Wong, and et~al., ``{BEHAVIOR Robot Suite: Streamlining Real-World Whole-Body Manipulation for Everyday Household Activities},'' \emph{Proc. 9th Conf. Robot Learning (CoRL)}, 2025.

\bibitem{RN5}
J.~Hu, P.~Stone, and R.~Martin-Martin, ``{Causal Policy Gradient for Whole-Body Mobile Manipulation},'' \emph{Proc. Robotics: Science and Systems (RSS)}, 2023.

\bibitem{RN6}
Z.~Fu, X.~Cheng, and D.~Pathak, ``{Deep Whole-Body Control: Learning a Unified Policy for Manipulation and Locomotion},'' \emph{Proc. 6th Conf. Robot Learning (CoRL)}, pp. 138--149, 2023.

\bibitem{RN7}
M.~Tjandrasuwita, J.~Xu, A.~Solar-Lezama, and W.~Matusik, ``{MeMo: Meaningful, Modular Controllers via Noise Injection},'' \emph{Proc. Advances in Neural Information Processing Systems (NeurIPS)}, 2024.

\bibitem{RN8}
J.-P. Sleiman, F.~Farshidian, M.~V. Minniti, and M.~Hutter, ``{A Unified MPC Framework for Whole-Body Dynamic Locomotion and Manipulation},'' \emph{IEEE Robot. Autom. Lett.}, vol.~6, no.~3, pp. 4688--4695, 2021.

\bibitem{RN9}
Z.~He and et~al., ``{PAKE: Learning Whole-Body Loco-Manipulation with Partial Kinematic Embeddings},'' \emph{arXiv preprint arXiv:2607.11041}, 2026.

\bibitem{RN10}
A.~Gupta, V.~Kumar, C.~Lynch, S.~Levine, and K.~Hausman, ``{Relay Policy Learning: Solving Long-Horizon Tasks via Imitation and Reinforcement Learning},'' \emph{Proc. 3rd Conf. Robot Learning (CoRL)}, pp. 875--893, 2020.

\bibitem{RN11}
B.~Ichter and et~al., ``{Do as I Can, Not as I Say: Grounding Language in Robotic Affordances},'' \emph{Proc. 6th Conf. Robot Learning (CoRL)}, pp. 287--318, 2022.

\bibitem{RN12}
H.~Liu and et~al., ``{RoboMatch: A Unified Mobile-Manipulation Teleoperation Platform with Auto-Matching Network Architecture for Long-Horizon Tasks},'' \emph{arXiv preprint arXiv:2509.08522}, 2025.

\bibitem{RN13}
C.~Yu and P.~Wang, ``{Dexterous Manipulation for Multi-Fingered Robotic Hands with Reinforcement Learning: A Review},'' \emph{Front. Neurorobot.}, vol.~16, p. 861825, 2022.

\bibitem{RN14}
J.~Egli, B.~Forrai, T.~Buchner, and et~al., ``{Sensorized Soft Skin for Dexterous Robotic Hands},'' \emph{Proc. IEEE Int. Conf. Robotics and Automation (ICRA)}, pp. 18\,127--18\,133, 2024.

\bibitem{RN15}
H.~Xue, J.~Ren, W.~Chen, and et~al., ``{Reactive Diffusion Policy: Slow-Fast Visual-Tactile Policy Learning for Contact-Rich Manipulation},'' \emph{Proc. Robotics: Science and Systems (RSS)}, 2025.

\bibitem{RN16}
M.~Murooka, T.~Hoshi, K.~Fukumitsu, and et~al., ``{TACT: Humanoid Whole-Body Contact Manipulation Through Deep Imitation Learning With Tactile Modality},'' \emph{IEEE Robot. Autom. Lett.}, vol.~10, no.~8, pp. 7819--7826, 2025.

\bibitem{RN17}
Y.~Liu, J.~Zhang, Y.~Lou, B.~Zhang, J.~Zhou, and J.~Chen, ``{Soft Bionic Gripper with Tactile Sensing and Slip Detection for Damage-Free Grasping of Fragile Fruits and Vegetables},'' \emph{Comput. Electron. Agric.}, vol. 220, p. 108904, 2024.

\bibitem{RN18}
A.~Iyer, Z.~Peng, Y.~Dai, and et~al., ``{OPEN TEACH: A Versatile Teleoperation System for Robotic Manipulation},'' \emph{Proc. 8th Conf. Robot Learning (CoRL)}, pp. 2372--2395, 2025.

\bibitem{RN19}
X.~Cheng, J.~Li, S.~Yang, G.~Yang, and X.~Wang, ``{Open-TeleVision: Teleoperation with Immersive Active Visual Feedback},'' \emph{Proc. 8th Conf. Robot Learning (CoRL)}, 2024.

\bibitem{RN20}
Z.~Fu, T.~Z. Zhao, and C.~Finn, ``{Mobile ALOHA: Learning Bimanual Mobile Manipulation Using Low-Cost Whole-Body Teleoperation},'' \emph{Proc. 8th Conf. Robot Learning (CoRL)}, 2024.

\bibitem{RN21}
Y.~Jiang and et~al., ``{BEHAVIOR Robot Suite: Streamlining Real-World Whole-Body Manipulation for Everyday Household Activities},'' \emph{Proc. 9th Conf. Robot Learning (CoRL)}, pp. 1246--1281, 2025.

\bibitem{RN22}
Y.~Qin, W.~Yang, B.~Huang, and et~al., ``{AnyTeleop: A General Vision-Based Dexterous Robot Arm-Hand Teleoperation System},'' \emph{Proc. Robotics: Science and Systems (RSS)}, 2023.

\bibitem{RN23}
Z.~Zhang, J.~Pang, Z.~Yang, and et~al., ``{Dexora: Open-Source VLA for High-DOF Bimanual Dexterity},'' \emph{arXiv preprint arXiv:2605.18722}, 2026.

\bibitem{RN24}
Z.-H. Yin, B.~Huang, Y.~Qin, Q.~Chen, and X.~Wang, ``{Rotating Without Seeing: Towards In-hand Dexterity Through Touch},'' \emph{Proc. Robotics: Science and Systems (RSS)}, 2023.

\bibitem{RN25}
I.~Güzey, B.~Evans, S.~Chintala, and L.~Pinto, ``{Dexterity from Touch: Self-Supervised Pre-Training of Tactile Representations with Robotic Play},'' \emph{Proc. 7th Conf. Robot Learning (CoRL)}, pp. 3142--3166, 2023.

\bibitem{RN26}
H.~Qi, B.~Yi, S.~Suresh, and et~al., ``{General In-hand Object Rotation with Vision and Touch},'' \emph{Proc. 7th Conf. Robot Learning (CoRL)}, pp. 2549--2564, 2023.

\bibitem{RN27}
L.~Heng, H.~Geng, K.~Zhang, P.~Abbeel, and J.~Malik, ``{ViTacFormer: Learning Cross-Modal Representation for Visuo-Tactile Dexterous Manipulation},'' \emph{arXiv preprint arXiv:2506.15953}, 2025.

\bibitem{RN28}
B.~Huang, Y.~Wang, X.~Yang, Y.~Luo, and Y.~Li, ``{3D-ViTac: Learning Fine-Grained Manipulation with Visuo-Tactile Sensing},'' \emph{Proc. 8th Conf. Robot Learning (CoRL)}, pp. 2557--2578, 2025.

\bibitem{RN29}
Y.~Chen, M.~van~der Merwe, A.~Sipos, and N.~Fazeli, ``{Visuo-Tactile Transformers for Manipulation},'' \emph{Proc. 6th Conf. Robot Learning (CoRL)}, pp. 2026--2040, 2023.

\bibitem{RN30}
P.~Lin, Y.~Huang, J.~Ma, and et~al., ``{PP-Tac: Paper Picking Using Omni-directional Tactile Feedback in Dexterous Robotic Hands},'' \emph{Proc. Robotics: Science and Systems (RSS)}, 2025.

\bibitem{RN31}
H.~Kim, Y.~Ohmura, and Y.~Kuniyoshi, ``{Goal-Conditioned Dual-Action Imitation Learning for Dexterous Dual-Arm Robot Manipulation},'' \emph{IEEE Trans. Robot.}, vol.~40, pp. 2287--2305, 2024.

\bibitem{RN32}
J.~Gao, X.~Jin, F.~Krebs, N.~Jaquier, and T.~Asfour, ``{Bi-KVIL: Keypoints-Based Visual Imitation Learning of Bimanual Manipulation Tasks},'' \emph{Proc. IEEE Int. Conf. Robotics and Automation (ICRA)}, pp. 16\,850--16\,857, 2024.

\bibitem{RN33}
T.~Z. Zhao, V.~Kumar, S.~Levine, and C.~Finn, ``{Learning Fine-Grained Bimanual Manipulation with Low-Cost Hardware},'' \emph{arXiv preprint arXiv:2304.13705}, 2023.

\bibitem{RN34}
C.~Chi, Z.~Xu, S.~Feng, E.~Cousineau, Y.~Du, B.~Burchfiel, R.~Tedrake, and S.~Song, ``{Diffusion Policy: Visuomotor Policy Learning via Action Diffusion},'' \emph{Int. J. Robot. Res.}, vol.~44, no. 10-11, pp. 1684--1704, 2025.

\bibitem{RN35}
S.~Lee, Y.~Wang, H.~Etukuru, and et~al., ``{Behavior Generation with Latent Actions},'' \emph{Proc. Int. Conf. Machine Learning (ICML)}, pp. 26\,991--27\,008, 2024.

\bibitem{RN36}
W.~Song and et~al., ``{ReconVLA: Reconstructive Vision-Language-Action Model as Effective Robot Perceiver},'' \emph{Proc. AAAI Conf. Artificial Intelligence (AAAI)}, 2026.

\bibitem{RN37}
T.~Z. Zhao and et~al., ``{ALOHA Unleashed: A Simple Recipe for Robot Dexterity},'' \emph{Proc. 8th Conf. Robot Learning (CoRL)}, pp. 1910--1924, 2025.

\bibitem{RN38}
A.~Prasad, K.~Lin, J.~Wu, L.~Zhou, and J.~Bohg, ``{Consistency Policy: Accelerated Visuomotor Policies via Consistency Distillation},'' \emph{Proc. Robotics: Science and Systems (RSS)}, 2024.

\bibitem{RN39}
A.~Mete, H.~Xue, A.~Wilcox, Y.~Chen, and A.~Garg, ``{QueST: Self-Supervised Skill Abstractions for Learning Continuous Control},'' \emph{Proc. Advances in Neural Information Processing Systems (NeurIPS)}, 2024.

\bibitem{RN40}
S.~Haldar, Z.~Peng, and L.~Pinto, ``{BAKU: An Efficient Transformer for Multi-Task Policy Learning},'' \emph{Proc. Advances in Neural Information Processing Systems (NeurIPS)}, pp. 141\,208--141\,239, 2024.

\bibitem{RN41}
A.~C.-W. Lee, I.~Chuang, L.-Y. Chen, and I.~Soltani, ``{InterACT: Inter-dependency Aware Action Chunking with Hierarchical Attention Transformers for Bimanual Manipulation},'' \emph{Proc. 8th Conf. Robot Learning (CoRL)}, pp. 1730--1743, 2025.

\bibitem{RN42}
B.~Wu, R.~Martin-Martin, and L.~Fei-Fei, ``{M-EMBER: Tackling Long-Horizon Mobile Manipulation via Factorized Domain Transfer},'' \emph{Proc. IEEE Int. Conf. Robotics and Automation (ICRA)}, pp. 11\,690--1169, 2023.

\end{thebibliography}

\end{document}